\documentclass[12pt]{article}
\usepackage[utf8]{inputenc}
\usepackage[english]{babel}

\usepackage{graphicx}
\usepackage{amsmath}
\usepackage{amssymb}
\usepackage{amsthm}
\usepackage[disable]{todonotes}
\usepackage[shortlabels]{enumitem}

\newtheorem{definition}{Definition}

\begin{document}
	\title{General Fragment Model for Information Artifacts}
	\author{Sandro Rama Fiorini, Wallas Sousa dos Santos, \\
	        Rodrigo Costa Mesquita, \\
	        Guilherme Ferreira Lima, Marcio F. Moreno \\
		IBM Research Brazil}
	%


	%
	\maketitle              % typeset the header of the contribution
	\begin{abstract}
		The use of semantic descriptions in data intensive domains require a systematic model for linking semantic descriptions with their manifestations in fragments of heterogeneous information and data objects. Such information heterogeneity requires a fragment model that is general enough to support the specification of anchors from conceptual models to multiple types of information artifacts. While diverse proposals of anchoring models exist in the literature, they are usually focused in audiovisual information. We propose a generalized fragment model that can be instantiated to different kinds of information artifacts. Our objective is to systematize the way in which fragments and anchors can be described in conceptual models, without committing to a specific vocabulary.  

	\end{abstract}

\section{Introduction}

The relationship between data and knowledge plays an important role in natural sciences. In Geology, for example, a big part of what constitutes geological knowledge is captured by its relationship with unstructured  data found in images, seismic cubes, well logs and tabular information. Systems for knowledge representation in these fields should be able to explicitly capture the relationship between knowledge and data.  This requirement  is often referred  as the semantic gap between symbolic models and sub-symbolic structures.

The semantic gap has been approached from different sides along the previous 15 years. However, the problem is still open. One important practical aspect is the need for a systematic way of referring to fragments of information. Consider a node of a knowledge graph representing a geological object in a complex geological model. Its existence can be linked to spatial regions in different seismic cubes, well log signals, core descriptions and so on. Representing this link requires us to be able to refer to and describe fragments of these data objects, which can then be retrieved and reasoned with.

Different works have addressed the problem of defining fragments on media objects, such as standards by ISO \cite{Amielh2006} and W3C \cite{Troncy2012}. However, to this day there has been no general way to specify fragments on any sort of data, with having clear and formal semantics.

In this  paper, we propose the General Fragment Model (GFM) to systematize the definition of fragments of generalized media types. It is based on the notion of indexers functions, that map arbitrary identification tokens to parts of information artifacts. It generalizes existing models, allowing arbitraty fragment schemas on not necessarily audiovisual media. We also discuss an implementation of GFM into the Hyperknowledge framework for knowledge representation, where media fragments can be specified and queried with a specific query language capable of exploiting the proposed model.

%- Querying of multimedia objects should have access to media fragments. Anchors are the mechanism used to tie descriptions to media fragments.
%
%- Increased need to link and query other types of information, which crosses the boundary of pure multimedia models. Users in specific domain want to query sensor data, ml models, and so on. More importantly, users want to link and query parts of these information artifacts.
%
%- While anchor models and ontology models for annotations exist, there is not much work trying to bring both together. Anchor models tend to be focused on audiovisual data, while annotation models might get very complicated.
%
%- We propose an anchor model that is flexible enough to allow reference and query of any kind of information artifact and is simple enough that it can be simply implemented. We define a conceptual model of fragments and a format to describe types of fragments and their instantiation.

\section{Related Work}

There are some published standards that specify hypermedia fragments. The MPEG Part 21 standard \cite{Amielh2006}  defines a URI-based fragment identification for media resources. It extends the W3C XPointer Framework Recommendation\footnote{https://www.w3.org/TR/xptr-framework/} to address spatiotemporal  and logical locations on resources, as well as tracks of media files, portions of video and byte ranges. Hausenblas \emph{et al.} \cite{Hausenblas2009} define a fragment schema  in four dimensions: times, space, track and names (or id). They are relatively simple, based on the standard format for queries with URIs. The two main dimensions (i.e. time and space) became the  W3C Recommendation Media Fragments URI 1.0\footnote{https://www.w3.org/TR/media-frags/} \cite{Troncy2012}. More recently, the EPUB Canonical Fragment Identifiers 1.1 \cite{Sorotokin2017} improved the generalization of fragment identification to include provisions for text fragments, along with time, space and content logic. These standards, while adequate for their requirements, present a too narrow approach to the definition of fragments. Considering the huge amount of heterogeneous, non-structured information present in the Web today, a more formal, general model for fragments is needed. The approach proposed in this paper generalize the previous contributions.

%Some works employ fragments and anchors in retrieval. Thomas \textit{et al.} \cite{Kurz2015} proposes a SPARQL engine with primitives to deal with media. It assumes a basic conceptual model of temporal and spatial entities, which can be used to define fragments on media content. It includes sparql primitives to deal with topological relations in space and time, directional relations, and aggregations over these.  

Some works go into a different direction, proposing ontological models to represent and query fragments. Thomas \textit{et al.} \cite{Kurz2015} proposes a SPARQL-based engine with primitives to deal with media, called SPARQL-MM. It assumes a basic conceptual model of temporal and spatial entities, which can be used to define fragments on media content. It includes SPARQL primitives to deal with topological relations in space and time, as well as directional relations (e.g. \emph{above}, \emph{right-of}), and aggregations over these (i.e. bounding box).  Similarly, Nimkanjana and Witosurapot \cite{Nimkanjana2018} propose an extension of SPARQL engine where temporal media fragments represented in RDF graphs can be queried with respect to their temporal relations. 

Our approach is somewhat similar to these, in that it proposes a conceptual model, reference format and a query mechanism for fragments. However, our formal model provides a generalized conceptual model for fragments, capable of indexing different types of fragments in distinct types of objects, while our proposed query mechanism for fragments can deal with more advanced constructions.

\section{Information objects}

In order to define a generalized notion of fragment, we need a generalized notion of information. Here, we introduce the concept of \emph{information artifact}. 

The nature of information has been already explored extensively in the literature. In particular, two different top-level ontologies formalize some notion of information body. In DOLCE ontology \cite{Gangemi2005}, an information object (e.g. The Lord of The Rings, by Tolkien) is an entity that is, among other things, realized by some medium (e.g. a particular exemplar of the book) and expresses some content (e.g. a propositional description of the story).  In BFO ontology \cite{Ceusters2015}, the concept information content entity represents a similar notion of an information entity that is about something.

Here, we take a simplified version of DOLCE's information object. An \emph{information artifact} is a codification of some propositional content that realized by some physical or virtual object. Examples of information artifacts are the text encoded in file, the image in photo, or the data in a table.  More generally, we consider information artifacts of all sorts, including image, video, sound, text, sensor, vectoral drawings, tables, programs, databases, ontologies, and so on. When we refer to information in the following text, we are referring always to information artifacts. 

Furthermore, we assume that information artifacts are finite. While it is possible to conceive entities such as infinite list of numbers, the necessity for a physical substrate restricts the applicability of this concept. Infinite information artifact are more akin to information generators, which we will not address here.

More importantly, information artifacts can be broken into parts. The concept is homeomeric: parts of information artifact are also information artifacts. However, the specific subtypes of information artifacts might not be  homeomeric (e.g. parts of a book are not a book). Moreover, we assume here that the part relation on information artifacts are transitive. 

Finally, we also assume that all possible information artifacts can be ultimately coded as a stream of bits. In that sense, each bit is equivalent to an information atom.

%Info object s can be potentially infinite. For example, if we state that the info object  N is a list of all natural numbers, we can define infinite fragments onto it. We also assume that any information artifact  can be coded in a (possible infinite) sequence of binary numbers, which we call the fundamental codification of the object.

\section{General Fragment Model}

\begin{figure*}
	\centering
	\includegraphics[width=1.0\linewidth]{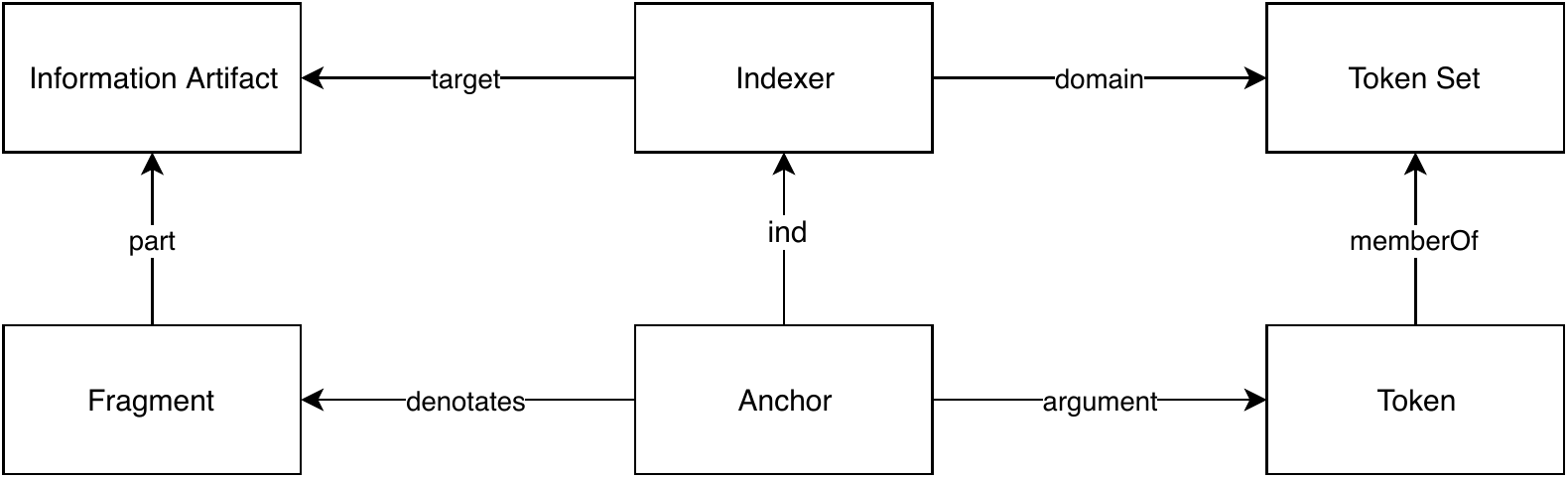}
	\caption{Conceptual model of the General Fragment Model.}
	\label{fig:anchormodel}
\end{figure*}

Our objective is to define a model to identify and situate parts within information artifacts. The \emph{General Fragment Model} (GFM) defines such model. It is composed by three main concepts: information artifact, indexer and anchor (Fig. \ref{fig:anchormodel}). In brief, an indexer defines a specific way of indexing (i.e. identifying) parts (or fragments) of a specific information artifact. It can be thought as a function that maps arbitrary tokens to parts of an information artifact. Each token identifies a part of an information artifact. An anchor is a particular token applied to a indexer targeted at an instantiated information artifact. 

More specifically, an (instance of an) \emph{indexer} is an abstract entity that describes a particular method of enumerating parts of a \emph{target} information artifact. An indexer is associated to a target information artifact, as well as with a \emph{token set}. A token set the types of tokens that are mapped to fragments of the information artifact. An anchor is then the application of an indexer with an specific token that denotes a part of an information artifact. The token set can be empty.

\begin{figure*}
	\centering
	\includegraphics[width=1.0\linewidth]{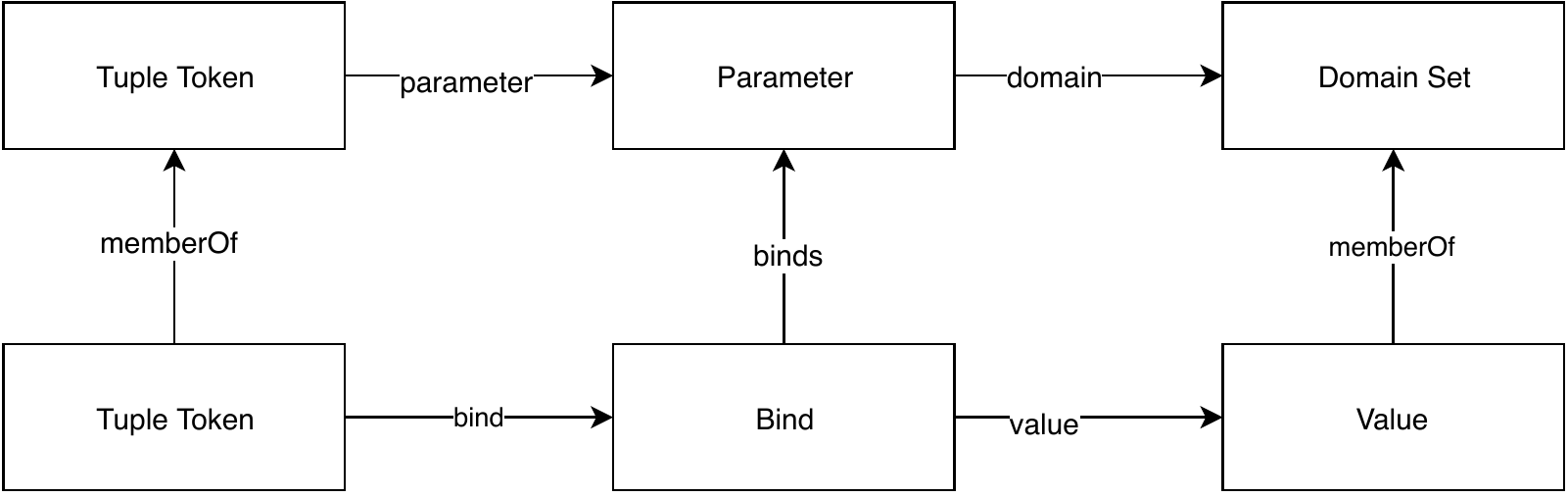}
	\caption{Model of tuple tokens.}
	\label{fig:model2}
\end{figure*}

\begin{figure*}
	\centering
	\includegraphics[width=1.0\linewidth]{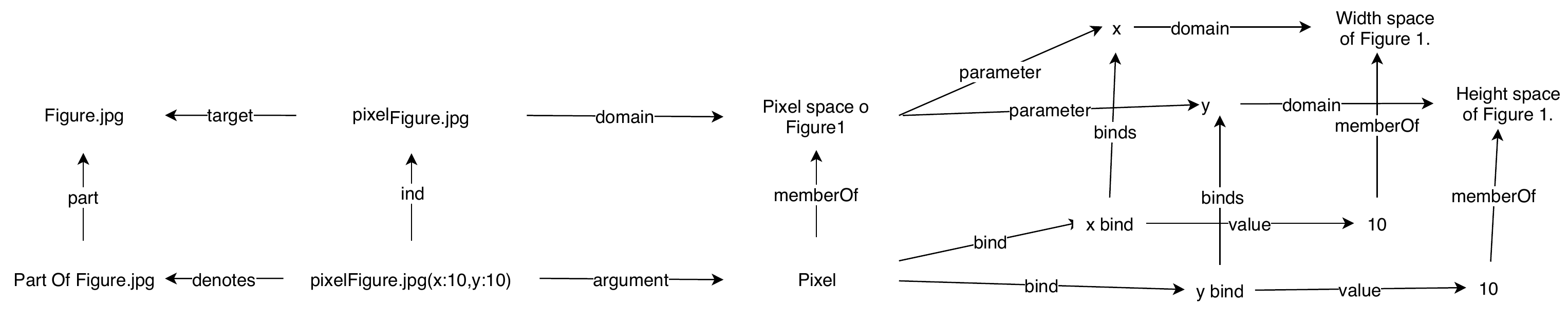}
	\caption{Instantiation of \emph{pixel} in GFM with tuple tokens. All labels are instances.}
	\label{fig:example1}
\end{figure*}

Tokens can be structured in many ways. We propose  a basic form which we call tuple token (\ref{fig:model2}). A tuple token is specified by a number of \emph{arguments} and \emph{domains}. An \emph{argument} denotes a set of elements of some kind which are used to compose an anchor. We call these sets \emph{domains}.  In this case, an anchor is an association of indexer with tuple token defined by binding a series of parameters to values in their domains. Consider, for instance, the \emph{pixel} fragment operator. It defines a pixel fragment on a target image, based on a two dimensional token. Fig. \ref{fig:example1} depicts an instantiation of GFM applying  \emph{pixel} to an image and defining an anchor on a specific pixel of that image. 

In the following text, we use the expression $f_o : p_1(D_1) \times \dots \times p_n(D_n) \to S$ to denote an indexer $f$ with target on an information artifact $o$ associated with a collection of parameter/domain pairs  $p_1(D_1), \dots , p_n(D_n)$ of a tuple token and an output type $S$. Similarly, we employ the expression $f_o[p_1(v_1), \dots , p_n(v_n)]=s$ to denote an anchor associated with an indexer $f$ on an object $o$ with a series of tuple token binds $p_1(v_1), \dots , p_n(v_n)$ and an output $s$. We can then specify a formal relation between an anchor and its index that limits the possible interpretations of GFM:
\begin{definition}
An anchor $a_o[r_1(v_1), \dots , r_n(v_n)]=s$ is an anchor of an indexer $f_o : p_1(D_1) \times \dots \times p_n(D_n) \to S$ iff $a_o = f_o$, $r_1 = p_1, \dots , r_n = p_n$ and $v_1 \in D_1, \dots ,v_n \in D_n$.
\end{definition}
We also assume that any information artifact has a binary indexer associated to it, which enumerates all bits of information of the information artifact:
\begin{definition}
For any information artifact $o$, there is a binary indexer $b_o : ind(N) \to B$, where $N \subset \mathbb{N}$ is the size in bits of $o$ and $B$ is the set of bits of $o$. 
\end{definition}

%An example of indexer is a rectangle selector on an image $I$ sized $N \times M$, which specified a function $rect: N^4 \to S$, where $S$ is the set of all subimages of  $I$. An anchor on $I$ is a bind $a \in N^4$, such that $rect(a)=s$, and $s$ is a subimage ot $i$. 

An important aspect of this model is that it does not assume any particular structure to information artifacts (apart from the fundamental binary codification). As a matter of fact, we can see indexers as the very mechanism that allows one to talk about parts on information artifacts.  For example, the fact that images are usually described in terms of pixels can be translated to the existence of a indexer that maps 2D coordinates to pieces of the image that can be described in terms of coordinates in the RGB domain.

%Given that an indexer can have arbitrary domains, any function can work as an indexer. Those range from trivial function such as the identity function, to more complex queries that take whole strings as inputs. 

Indexers can be combined in different ways. Indexers can be concatenated, defining indexers on indexers. There two possible meanings to this construction. First, it is possible to define an anchor on an anchor (i.e. fragment of a fragment). So, considering the temporal indexer ${time}_o : s(N) \times f(N) \to T_o$ where $o$ is an audio file of length $N$ and $T_o$ is the set of intervals of $o$ and an anchor ${time}_o[ p(10), f(15)]=t$ selecting a 5 seconds interval o $o$ denoted by $t$, we can instantiate another temporal indexer ${time}_t : s(N) \times f(N) \to T_t$ that allows one to define an sub-anchor ${time}_t[ p(0), f(2)]=k$ that selects a 2 seconds interval within the fragment $t$. Another interpretation is to apply different indexers in sequence over  multidimensional information. We can apply an indexer on a piece of text that selects the $i$-th paragraph and then a character indexer that selects the $k$-th char in that paragraph.

Furthermore, indexers parameters can take any set as indexer domain, even the output of another indexer. This allows us to use fragments of an information artifact as input to indexers on itself or other information artifacts. Consider an indexer \[{colormask} : {color}(N) \to P\] that selects all pixels with a given color in an image and an indexer \[{pbounding}: {pixels}(P) \to R\] that selects the bounding rectangle of a given set of pixels on the same image. We can compose these two indexers to form a complex anchor such that \[{pbounding}[ {pixel}({colormask}[color(\mathrm{'red'})])]\].

\subsection{Basic Indexer Taxonomy}

\begin{figure}
	\centering
	\includegraphics[width=0.8\linewidth]{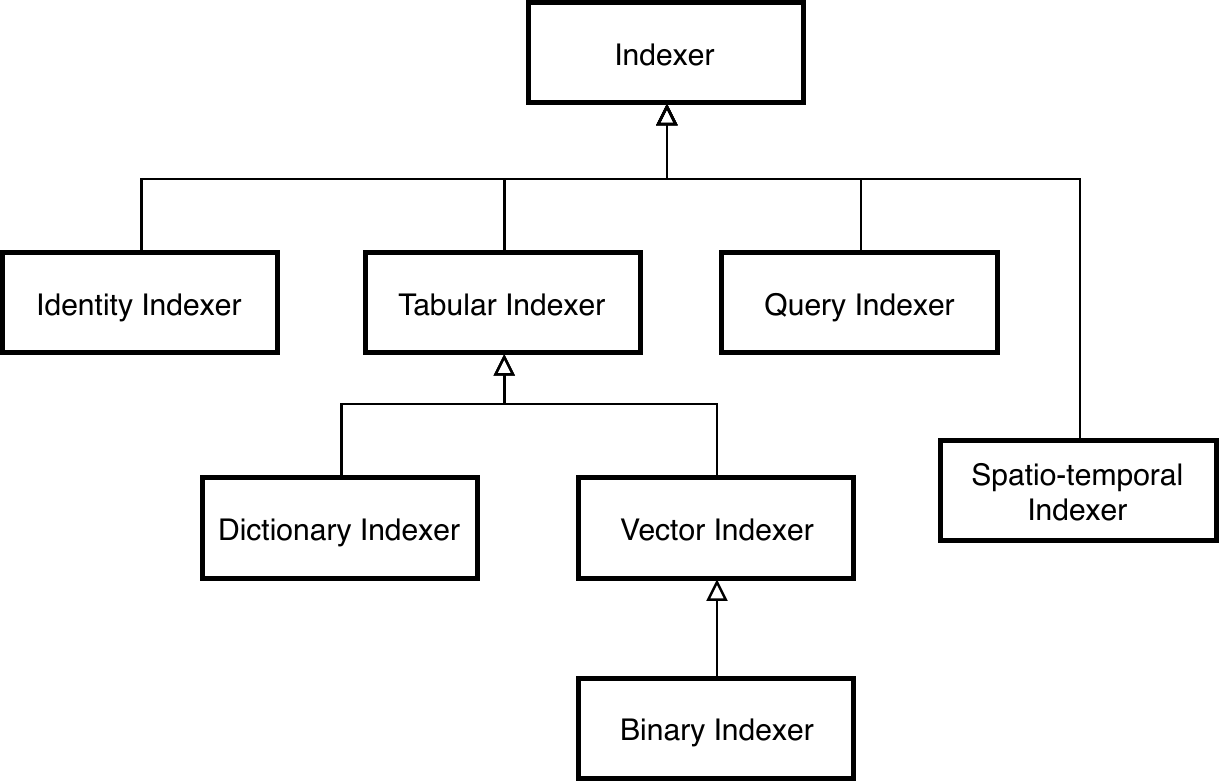}
	\caption{Indexer taxonomy.}
	\label{fig:indexertaxonomy}
\end{figure}

Taking the general anchor model, it is possible to specialize it with a basic taxonomy of indexer types (Fig. \ref{fig:indexertaxonomy}). In the following we present some basic, non-disjoint indexer categories. 

\begin{description}
	\item[Binary Indexer] The binary indexer is the most fundamental type of index and it is assumed to exist to any information artifact. A binary indexer denotes a function $binary : N \to S_B$, such that B is the set of all bits of the information artifact.
	\item[Identity indexer] The identity indexer on an object $R$ (with parts set $S$) is a function $id : S \to S$, such that $id(s)=s$ for all $s \in S$.
	\item[Tabular Indexer] A tabular indexer enumerates a disjoint set of parts on a information artifact, such that each possible anchor maps to a single part. There two main subtypes. Vector indexers impose a vectoral space on the information artifact, so that all indexer domains are totally ordered sets, usually numeric. They induce a partial-order of information parts (i.e. pixels on image, words on text, frames on a video). The binary indexer is also a type of vector indexer. On the other hand, dictionary indexers are tabular indexers that take non-ordered sets of symbols as inputs, such as strings, and mapped to individual parts of the information artifact. 
	\item[Spatio-temporal indexer] An indexer that impose a spatial view on the data, allowing selection of $n$-dimensional regions of the data. 
	\item[Query indexer] While all indexers can be seen as construed on the data, query indexers are more akin to prepared statements in query languages, such as SQL, where the indexer domains take query strings that are used to select elements of the information artifact. An instance of a query indexer is a specific query schema (or formula) applied to an information artifact. In effect, it allows one to refer to parts of an information artifact according to some criteria. This type of indexer is intended to be attached to information artifacts denoting ordered collections. It can be targeted to any resource, even query engines. For example, one can think of a query indexer targeted at Google is reifying the search engine as an information artifact formed by millions of individual document entries.
\end{description}

\section{Hyperknowledge Implementation}

The GA Model can potentially be instantiated in distinct representation languages. Here we present an implementation based on Hyperknowledge, a knowledge representation  framework focused on multimedia content. The implementation instantiates  GFM as an extension of Hyperknowledge`s anchor model.

%Another way of instantiating GFM is to use conceptual models and ontologies to represent indexers and anchors explicitly. This idea have been explored semantic annotation frameworks for multimedia and other information types, notably \cite{}, allowing for content-based query of media and media fragments. These frameworks are OWL/RDF-based, which are limited formalisms in terms of expressivity, without clear native support for anchors. In this section, we briefly explore the instantiation of GFM in Hyperknowledge, a knowledge representation framework particularly tailored for hypermedia representation. It is derived from the NCM [ref] model for ... \todo{Finish} . It incorporates anchors as one of its main constructs.

Hyperknowledge is derived from the NCM model for hypermedia document representation \cite{Moreno2017} . Hyperknowledge models incorporate media content directly into the knowledge model. In its original definition, Hyperknowledge allows the modeler to link graph descriptions to fragments of media using constructs called anchors. However, there is no clear semantics for this construct. Here we incorporate the GA Model into Hyperknowledge to address this issue.

Hyperknowledge has three basic constructs: \emph{nodes}, \emph{links} and \emph{anchors}. Nodes represent any sort of entity, from concepts to information artifacts. 
%Nodes can have subnodes, in which case they are called contexts. 
Links refer to $n$-ary relationships between nodes. Links are associated to nodes through anchors defined on nodes. Anchors represent a fragment of the content represented by a node. An anchor on a node denoting an image might represent a fragment of the image, or a individual pixel. For example, a spatial anchor on a image node may represent a sub region of an image. A temporal anchor represents a temporal segment (interval) of a continuous media (e.g., audio, video). Other types of anchors are possible. In order to simplify the definition of anchors in HK, we introduce a simplified definition of HK (meta) model :
\begin{definition}
A hyperknowledge model $M$ is a tuple $(N, L, P, A)$, where: (a) $N$ is a set of nodes; (b) $L$ is a set of links; (c) $P$ is a set of property-value pairs $(p, v)$; and (d) $A$ is a set of anchors. Let also $P(n\#a)$ denote the set of properties associated with an anchor $a$ on a node $n \in N$ and $A(n)$ denote the set of anchors of $n$.
\end{definition}

We also denote links as first-order logic predicates. For example, $between(x,y,z)$ denotes a ternary link associating the anchors $x$, $y$ and $z$. When links are defined on lambda anchors, we used  nodes directly for clarity.

Finally, based on this definition, an HK anchor can be defined as:
\begin{definition}
An hk-anchor $a \in A$ of an model $M$ is a tuple $a = (id, n, P)$, where $id$ is a string denoting its identifier; $n$ is the target node of $a$; and $P$ is a set of propety-value pairs characterizing $a$. For any node $n$, $A(n)$ denotes the set of anchors on $n$. Also, for any node $n$, the anchor $\lambda \in A(n)$ denotes the whole information content of $n$.
\end{definition}
The lambda anchor $\lambda$ and the node itself can be seen as equivalent, having the same referent. So, properties attributed to a node are attributed to its lambda node.  We might also denote  $a = (id, n, P)$ as $n\#a$ for simplicity.

\section{Final Remarks}

We presented the General Anchor Model, a general theory about fragments of information artifacts. It is based on the notion of an indexer function that maps anchors, or combinations of parameters, to parts of information artifacts. It generalizes previous models and implementations presented in the literature, providing a systematic basis for implementation of anchor and fragment support in Web-based technologies. We also discussed how GFM is implemented in Hyperknowledge, a framework for knowledge representation on media contents. Particularly, we show how it can also benefit querying as well as representation. For future work, we plan to further explore the IRI-based language for anchor specification as well as generalize the indexer-anchor format to represent data processing mechanisms.

%
% The next two lines define the bibliography style to be used, and the bibliography file.
%\bibliographystyle{plain}
%\bibliography{bibs}

\end{document}